\documentclass[runningheads]{llncs}
\usepackage{graphicx}
\usepackage{comment}
\usepackage{booktabs}
\usepackage{amsmath,amssymb} %
\usepackage{color}
\usepackage{multirow}
\usepackage{wrapfig}
\usepackage{xcolor}
\usepackage{cite}
\usepackage{isomath} %

\newcommand\etal{\textit{et~al.}}
\newcommand\ie{\textit{i.e.}}
\newcommand\eg{\textit{e.g.}}

\newcommand{\argmin}{\operatornamewithlimits{argmin}}

\definecolor{citecolor}{HTML}{0071bc}
\usepackage[pagebackref=true,breaklinks=true,letterpaper=true,colorlinks,citecolor=citecolor,bookmarks=false]{hyperref}

\begin{document}
\newcommand\blfootnote[1]{%
  \begingroup
  \renewcommand\thefootnote{}\footnote{\hspace{-7pt}#1}%
  \addtocounter{footnote}{-1}%
  \endgroup
}
\pagestyle{headings}
\mainmatter
\def\ECCVSubNumber{1148}  %

\title{Label-Efficient Learning on Point Clouds using Approximate Convex Decompositions
} %

\titlerunning{Label-Efficient Learning on Point Clouds using ACD}
\author{Matheus Gadelha$^*$, 
Aruni RoyChowdhury$^{*\ddagger}$, 
Gopal Sharma,
Evangelos Kalogerakis, 
Liangliang Cao, 
Erik Learned-Miller, 
Rui Wang, 
Subhransu Maji}
\authorrunning{Gadelha, RoyChowdhury et al.}
\institute{University of Massachusetts Amherst\\
\email{\{mgadelha,aruni,gopal,kalo,llcao,elm,ruiwang,smaji\}@cs.umass.edu}}
\maketitle

\begin{abstract}
The problems of shape classification and part segmentation from 3D
point clouds have garnered increasing attention in the last few
years. %
Both of these problems, however, suffer from relatively small training sets, creating the need for statistically efficient methods to learn 3D shape representations. 
In this paper, 
we investigate the use of Approximate Convex Decompositions (ACD) as a self-supervisory signal for label-efficient learning of point cloud representations.
We show that using ACD to approximate ground truth segmentation provides excellent self-supervision for learning 3D point cloud representations that are highly effective on downstream tasks. 
We report improvements over the state-of-the-art for unsupervised representation learning on the ModelNet40 
shape classification dataset and significant gains in few-shot part segmentation on the ShapeNetPart dataset.
Our source code is publicly available.\footnote{\url{https://github.com/matheusgadelha/PointCloudLearningACD} }

\blfootnote{$^*$ equal contribution.}
\blfootnote{$^{\ddagger}$ Now at Amazon, work done prior to joining.}

\end{abstract}

\section{Introduction}
\begin{figure}[t]
    \centering
    \includegraphics[width=\linewidth]{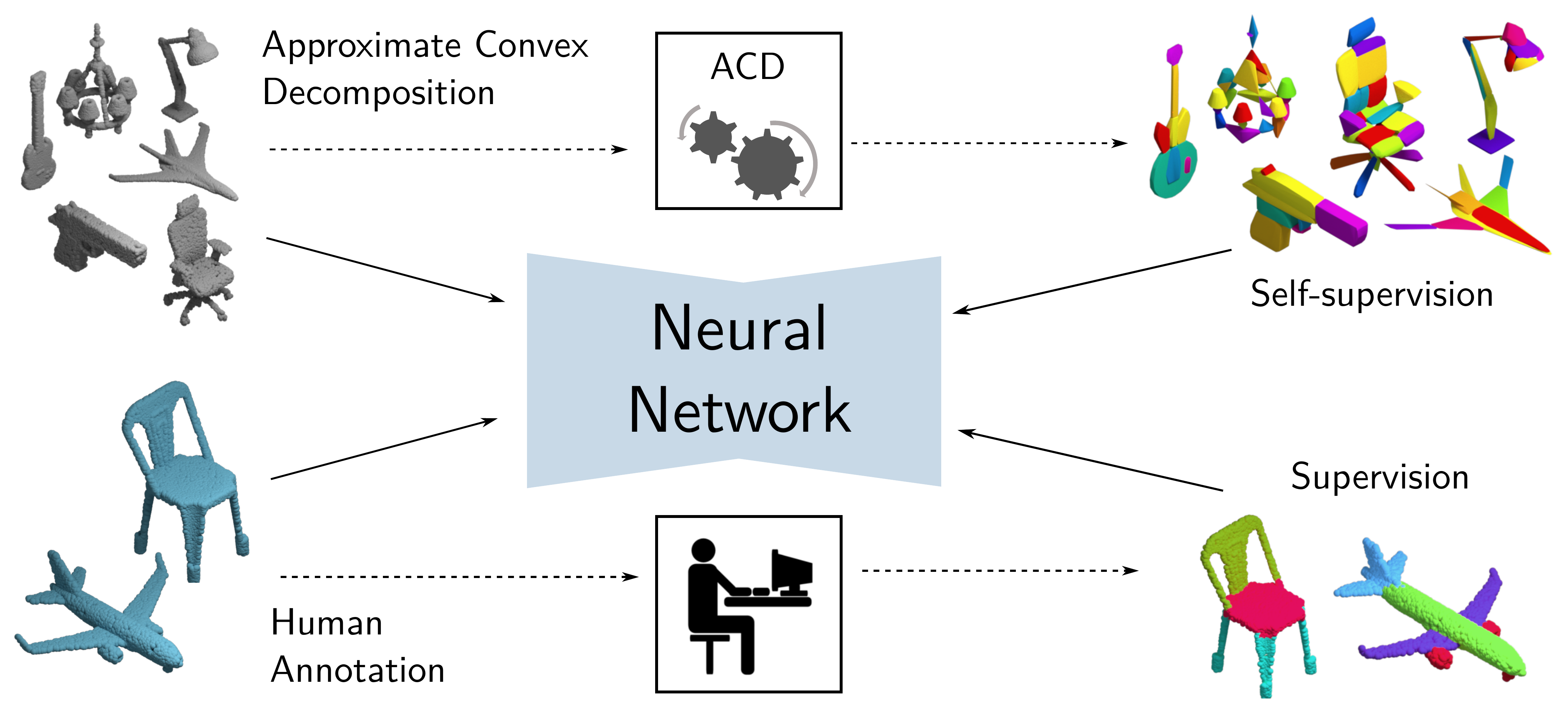}
    \caption{\small{Overview of our method versus a fully-supervised approach. \textit{\textbf{Top:}} Approximate Convex Decomposition (ACD) can be applied on a large repository of unlabeled point clouds, yielding a \textit{self-supervised} training signal for the neural network without involving any human annotators. \textit{\textbf{Bottom:}} the usual \textit{fully-supervised} setting, where human annotators label the semantic parts of shapes, which are then used as supervision for the neural network. The unsupervised ACD task results in learning useful representations from unlabeled data, significantly improving performance in shape classification and semantic segmentation, especially when labeled data is scarce or unavailable.}
    }
    \label{fig:vis_summary}
\end{figure}

The performance of current neural network models on tasks such as classification and semantic segmentation of point cloud data is limited 
by the
amount of labeled training data. 
Since collecting high quality annotations on 3D shapes is usually expensive and time consuming, there have
been increasing efforts to train on noisy or weakly labeled datasets~\cite{Muralikrishnan17,cosegnetChenyang,SharmaPEN},
or via completely unsupervised training~\cite{hassani2019unsupervised,mrt18,yang2018foldingnet,yang2019pointflow,chen2019bae}. 
An alternative strategy is to train the network on one task through {\em self-supervision} with automatically generated labels to initialize its parameters, then fine-tune it on the final task.
Examples of such self-supervised approaches for network initialization include clustering~\cite{caron2018deep,caron2019unsupervised}, solving jigsaw puzzles~\cite{noroozi2016unsupervised},  and image colorization~\cite{zhang2016colorful}. 
A key question in our problem setting is \textit{``What makes for a good self-supervision task in the case of 3D shapes?''} That is, what tasks induce inductive biases that are beneficial to the downstream
shape understanding tasks. 

\begin{figure}[t]
    \centering
    \vspace{2mm}
    \includegraphics[width=\linewidth]{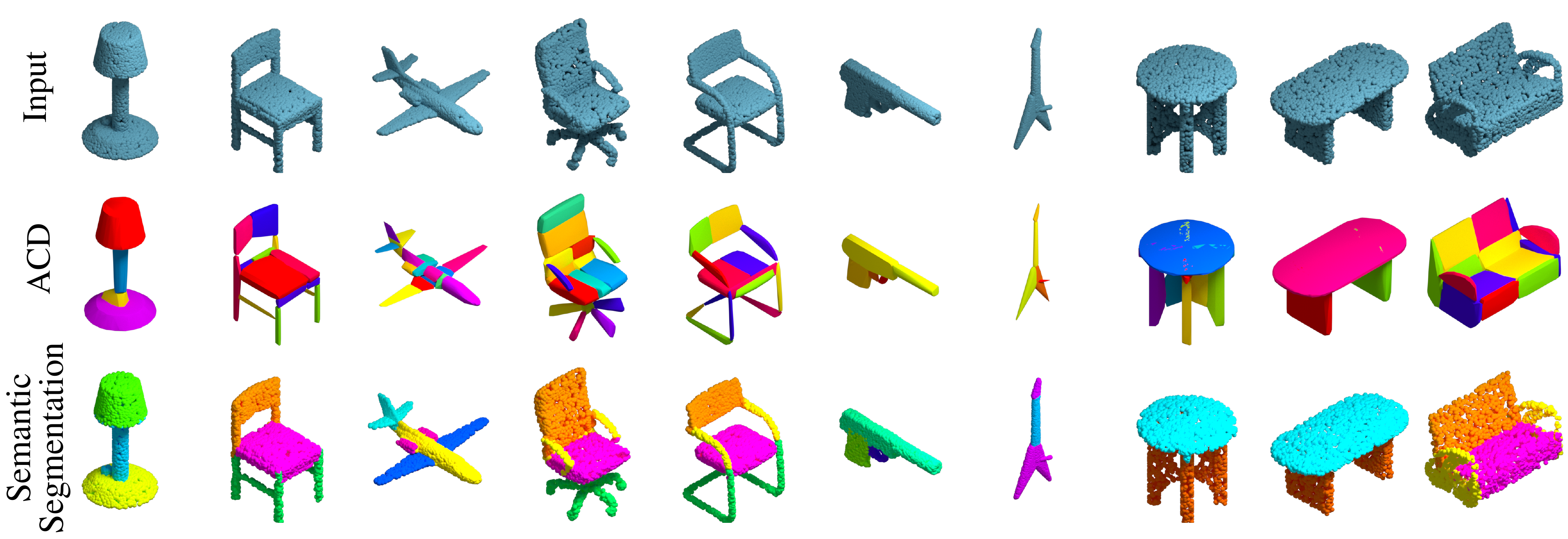}
    \vspace{2mm}
    \caption{ \small{
    Input point clouds (first row), convex components \emph{automatically} computed by ACD (second row) and
    \emph{human-labeled} point clouds (last row) from the ShapeNet~\cite{Chang2015ShapeNetAI} part segmentation benchmark.
    Note -- \textit{\textbf{(i)}} different colors for the ACD components only signify different parts-- no semantic
    meaning or inter-shape correspondence is inferred by this procedure; 
    \textit{\textbf{(ii)}} for the human labels, colors do convey semantic meaning: \eg, the backs of chairs are always orange;
    \textit{\textbf{(iii)}} while the ACD decompositions tend to oversegment the shapes, they contain 
    most of the boundaries present in the human annotations, 
    suggesting that the model has similar criteria for decomposing objects into subparts;
    \eg, the chair's legs are separated from the seat,
    wings and engines are separated from the airplane boundary, pistol trigger is separated from the rest.
    }}
    \label{fig:samples-of-acds}
\end{figure}

We posit that decomposing shapes into geometrically
simple constituent parts provides an excellent self-supervisory learning
signal for downstream 3D segmentation and classification tasks. 
Specifically, we propose using a classical shape decomposition method, Approximate Convex Decomposition (ACD), as the
self-supervisory signal to pre-train neural networks for processing 3D
data. Our approach is motivated by the observation that convexity provides cues to capture structural components
related to  human perception \cite{acd,acanalysis}, and object manufacturability \cite{Luo12,Demir18}.
However, strict convex decomposition often leads to highly
over-segmented shapes, leading to our choice of \textit{approximate}
convex decomposition \cite{acd}. 
As shown in the Figure
\ref{fig:samples-of-acds}, ACD decomposes shapes into segments that
roughly align with instances of different parts. For example, two wings
of an airplane are decomposed into two separate, approximately convex
parts. 

Our approach is illustrated in Figure~\ref{fig:vis_summary}.  The main
idea is to automatically generate training data by decomposing
unlabeled 3D shapes into convex components.  Since ACD relies solely
on geometric information to perform its decomposition, the process
does not require any human intervention.
We formulate ACD as a metric learning problem on point embeddings and
train the model using a contrastive
loss~\cite{hadsell2006dimensionality,chopra2005learning}.  
We demonstrate the
effectiveness of our approach on standard 3D shape classification and
segmentation benchmarks.  In classification, we show that the
representation learned from performing shape decomposition leads to
features that achieve state-of-the-art performance for \emph{unsupervised shape classification} on
ModelNet40~\cite{wu20153d} ($\mathbf{89.8\%}$).  For
\emph{few-shot shape segmentation} on ShapeNet~\cite{Chang2015ShapeNetAI}, our
model outperforms the state-of-the-art by \textbf{7.5\%} mIoU when
using 1\% of the available labeled training data.  Moreover, differently from
other unsupervised approaches, our method can be applied to any of the
well-known neural network backbones for point cloud processing.
Finally, we provide thorough experimental analysis and visualizations
demonstrating the role of the ACD self-supervision on 
representations learned by neural networks.

\section{Related Work}

\noindent\textbf{Learning Representations on 3D data.}
Shape representations using neural networks have been widely studied in computer vision and graphics.
An early approach was the use of \textit{occupancy grids} to represent shapes for classification or segmentation
tasks~\cite{wu20153d,maturana2015voxnets}; however these representations suffered from computational and memory inefficiency. These problems were mitigated by architectures that use spatial partitioning data structures~\cite{riegler2017octnet,klokov2017escape,wang2017ocnn,Wang:2018:AOP}.
\textit{Multi-view approaches}~\cite{Huang:2017:LMVCNN,Tat2018,kalogerakis2017shapepfcn,su15mvcnn,su2018deeper} learn representations 
by using order invariant pooling of features from multiple rendered views of a shape. 
Another class of methods take \textit{point cloud representations} (\ie, a set of $(x, y, z)$ coordinate triples) as input, and learn permutation 
invariant representations~\cite{wang2019dynamic,mrt18,qi2017pointnet,qi2017pointnetpp,yang2019pointflow,hassani2019unsupervised,pcagan,su18splatnet}. 
Using point sets as a 3D representation does not suffer from the memory constraints
of volumetric representations nor the self-occlusion issues of multi-view approaches.
Still, all the above approaches rely on massive amounts of labeled 3D data. 
In this paper, we develop a technique to enable the learning of  label-efficient  representations from point clouds.
Our approach is architecture-agnostic and relies on learning from approximate convex decompositions, which
can be automatically computed from a 3D shapes.

\vspace{2mm}
\noindent\textbf{Approximate Convex Decompositions.} 
Studies in the cognitive science literature have argued that humans tend to reason about 3D shapes as the union
of convex components~\cite{hoffman1983parts}. 
However, performing exact convex decomposition is an NP-Hard problem that leads to an impractically high number of
shape components~\cite{ecd}.
An alternative class of decomposition techniques, named \emph{Approximate Convex Decomposition} (ACD)~\cite{acd,minimumncd,acanalysis,vhacd}, compute approximate decompositions
up to a concavity tolerance $\epsilon$.
This tolerance makes the computation significantly more efficient and leads to shape approximations containing
a smaller number of components.
Apart from shape segmentation ~\cite{acanalysis,concavityaware}, 
these approximations are useful for a variety of tasks like mesh generation~\cite{acd} and collision detection~\cite{collisiondetection}.
Furthermore, ACD-based decompositions has been shown to have a high degree of similarity to human segmentations, as indicated by segmentation evaluation measures, such as the Rand Index in the PSB benchmark~\cite{Chen:2009:ABF,acanalysis}. This indicates that ACD can provide useful cues for discovering shape parts.

There have been various techniques to compute ACD for shapes based on either geometric analysis ~\cite{acd,minimumncd,acanalysis,vhacd}, and recently with deep networks \cite{deng2019cvxnet, chen2020bspnet}.
In this work, we used a particular type of ACD named Volumetric Hierachical Approximate Convex Decomposition (V-HACD)~\cite{vhacd} (more details in Section~\ref{sec:acd}.)
Unlike previous methods, our goal is to apply ACD to automatically supervise a convex decomposition task as an initialization step for learning point cloud representations.
We show that the training signal provided by ACD leads to improvements in semantic segmentation as well as
unsupervised shape classification.

\noindent\textbf{Self-supervised learning.}
In many situations, unlabeled images or videos contain useful information that can be leveraged to automatically create a training loss for learning useful representations. \textit{Self-supervised learning} explores this idea, using unlabeled data to train deep networks by solving  tasks that do not require any human annotation effort. 

Learning to colorize grayscale images %
was among the first  approaches to training modern
deep neural networks in a self-supervised fashion~\cite{larsson2016learning,zhang2016colorful,zhang2017split}. Estimating the color for a pixel from a black-and-white image requires some understanding of 
a pixel's meaning (\eg, skies are blue, grass is green, etc.). Thus training  a network to estimate pixel colors leads to 
the learning of representations that are useful in downstream tasks like object classification. 
The contextual information in an image also lends itself to the design of proxy 
tasks. These include learning to estimate the relative positions of cropped image patches~\cite{doersch2015unsupervised}, the similarity of patches tracked 
across videos~\cite{wang2015unsupervised,wang2017transitive}, the appearance of
missing patches in an image~\cite{pathak2016context,trinh2019selfie}, or 
learning from image modalities in RGBD data~\cite{gupta2016cross,su2018deeper}. 
Motion from unlabeled videos 
also provides a useful pre-training signal. Pathak~\etal~\cite{pathak2017learning} used motion segmentation to learn how to segment images, and
Jiang~\etal~\cite{jiang2018self} estimate relative depth as a proxy task for pre-training a network for scene understanding tasks. 
Other approaches include solving jigsaw puzzles 
with permuted image patches~\cite{noroozi2016unsupervised}, and training a 
generative adversarial model~\cite{donahue2019large}. An empirical comparison of 
various self-supervised tasks may be found in \cite{goyal2019scaling,kolesnikov2019revisiting}. 
In the case of limited samples, \ie, the \textit{few-shot classification} setting,
including self-supervised losses along with the usual supervised training is shown 
to be beneficial~\cite{su2019does}. 
Recent work has also focused on learning unsupervised representations for 
3D shapes using tasks such as clustering~\cite{hassani2019unsupervised} and 
reconstruction~\cite{sharma2016vconv,yang2018foldingnet}, which we  
compare against in our experiments.

\noindent\textbf{Label-efficient representation learning on point clouds.}
Several recent approaches \cite{Muralikrishnan18,cosegnetChenyang,hassani2019unsupervised,chen2019bae,SharmaPEN} have been proposed to
alleviate expensive labeling of shapes. Muralikrishnan \etal~\cite{Muralikrishnan18} learn a per-point representation by training a
network to predict shape-level tags. Yi et al.~\cite{Yi:2017:LHS}
embed pre-segmented parts in descriptor space by jointly learning a
metric for clustering parts, assigning tags to them, and building a
consistent part hierarchy. 
Chen \etal~\cite{chen2019bae} proposed a branched auto-encoder, where each branch
learns coarse part level features, which are further used to
reconstruct the shape by producing implicit fields for each part.
However, this approach requires one decoder for every different part, which
restricts their experiments to category-specific models.
On the other hand, our approach can be directly applied to any of the well known
point-based architectures, is capable of handling multiple categories at once
for part segmentation, and additionally learns useful features for unsupervised shape classification.
Furthermore, Chen \etal~\cite{chen2019bae} show experiments on single shot semantic segmentation on manually
selected shapes, whereas we show results on randomly selected training
shapes in a few-shot setting.
Most similar to our work, Hassani \etal~\cite{hassani2019unsupervised} propose a novel architecture for
point clouds, which is trained on multiple tasks at the same time: clustering, classification and reconstruction.
In our experiments, we demonstrate that we outperform their method on few-shot segmentation by $\mathbf{7.5\%}$ IoU and
achieve the same performance on unsupervised ModelNet40 classification by using \emph{only ACD as a proxy task}.
If we further add a reconstruction term, our method achieves state-of-the-art performance in unsupervised shape classification.
Finally, Sharma \etal~\cite{SharmaPEN} proposed learning point embeddings by using noisy
part labels and semantic tags available on the 3DWarehouse
dataset \cite{warehouse}. Their model is used for few-shot semantic
segmentation. 
In this work, we instead gather part labels using approximate convex decomposition, whose computation is automatic and can be applied to any mesh regardless of the existence of semantic tags.

\section{Method}

\subsection{Approximate Convex Decomposition}
\label{sec:acd}
In this subsection, we provide an overview of the shape decomposition approach used to 
generate the training data for our self-supervised task.
A detailed description of the method used in this work can be found in~\cite{vhacd}.

Given a polyhedron $P$, the goal is to compute the smallest set of convex polyhedra $\mathcal{C} = \{C_k | k=1,...,K\}$, 
such that the union $\cup_{k=1}^{K}C_i$ corresponds to $P$.
Exact convex decomposition of polyhedra is an NP-Hard problem~\cite{ecd} and 
leads to decompositions containing too many components, rendering it impractical for use
in most applications (ours included).
This can be circumvented by Approximate Convex Decomposition (ACD) techniques.
ACD relaxes the convexity constraint of exact convex decomposition by allowing every
component to be \emph{approximately convex} up to a concavity $\epsilon$.
The way concavity is computed and how the components are split varies according to different 
methods~\cite{vhacd,fastacd,acd,acanalysis,minimumncd}.
In this work, we use an approach called Volumetric Hierarchical Approximate Convex Decomposition (V-HACD)~\cite{vhacd}.
The reasons for using this approach are threefold.
First, as the name suggests, V-HACD performs computations using volumetric representations, which can be easily
computed from dense point cloud data or meshes and lead to good results
without having to resort to costly mesh decimation and remeshing procedures.
Second, the procedure is reasonably fast and can be applied to open surfaces of arbitrary genus.
Third, V-HACD guarantees that no two components overlap, which means that there is no
part of the surface that is approximated by more than one component.
In the next paragraph, we describe V-HACD in detail.

\vspace{2mm}
\noindent{\textbf{V-HACD.}} Since the method operates on volumetric representations,
the first step is to convert a shape into an occupancy grid.
If the shape is represented as a point cloud, one can compute an occupancy grid by
selecting which cells are occupied by the points and filling its interior.
In our case, since our training shapes are from ShapeNet~\cite{Chang2015ShapeNetAI} that includes meshes, we chose to compute the occupancy grid by voxelizing the meshes using~\cite{voxelizer}.
Once the voxelization is computed, the algorithm proceeds on computing convex components by
recursively splitting the volume into two parts.
First, the volume is centered and aligned in the coordinate system according to its principal
axis.
Then, one of the three axis aligned planes is selected as a splitting plane that
separates the volume in two different parts.
This procedure is applied multiple times until we reach the maximum number of desired components
or the concavity tolerance is reached.
The concavity $\eta(\mathcal{C})$ of a set of components $\mathcal{C}$ is computed as 
\begin{equation}
\eta(\mathcal{C}) = \max_{k=1,...K}d(C_k, \mathtt{CH}(C_k)),
\end{equation}
where $d(X,Y)$ is the difference between the volumes $X$ and $Y$;
$\mathtt{CH}(X)$ is the convex hull of $X$;
and $C_k$ is the $k$th element of the set $\mathcal{C}$.
The splitting plane selection is performed by choosing the axis-aligned plane that minimizes an energy $E(V, \mathbf{p})$,
where $V$ is the volume we are aiming to split and $\mathbf{p}$ is the splitting plane.
This energy is defined as:
\begin{equation}
    E(V, \mathbf{p}) = E_{con}(V, \mathbf{p}) + \alpha E_{bal}(V, \mathbf{p}) + \beta E_{sym}(V, \mathbf{p}),
\end{equation}
where $E_{con}$ is a component connectivity term, which measures the sum of the normalized concavities between both sides of volume;
$E_{bal}$ is a balance component term, which measures the dissimilarity between both sides;
and $E_{sym}$ is a symmetry component term, which penalizes planes that are orthogonal to a potential revolution axis.
The parameters $\alpha$ and $\beta$ are weights for the last two terms.
In all our experiments, we used the default values of $\alpha = \beta = 0.05$.
We refer the reader to~\cite{vhacd} for a detailed description of the components in the energy term.

\vspace{2mm}
\noindent{\textbf{Assigning component labels to point clouds.}} 
The output of ACD for every shape is a set of approximately convex components represented as meshes.
For each shape, we sample points on the original ShapeNet mesh and on the mesh of every ACD component.
We then propagate component labels to every point in the original point cloud by using
nearest neighbor matching with points used in the decomposition. 
More precisely, given an unlabeled point cloud $\{ p_i \}_{i=1}^{N}$, this assigns a component 
label $\mathrm{\Gamma}(p_i, \mathcal{C})$ to each point $p_i$ via:

\begin{equation}
    \mathrm{\Gamma}(p_i, \mathcal{C}) = \argmin_{k = 1 \dots |\mathcal{C}|} \bigg[ \min_{p_j \in C_k} || p_i - p_j ||  \bigg].
\end{equation}

\subsection{Self-supervision with ACD}
 The component labels generated by the ACD algorithm are not consistent across point clouds, 
 \ie, ``component 5'' may refer to the \textit{seat} of a chair in one point cloud 
 but the \textit{leg} of the chair in another. 
 Therefore, the usual cross-entropy loss, which is generally used to train networks for tasks such as semantic part labeling, 
 is not applicable in our setting. We formulate the learning of ACDs
 as a metric learning problem on point embeddings via a \textit{pairwise} or \textit{contrastive loss}~\cite{hadsell2006dimensionality}.

We assume that each point $p_i = (x_i, y_i, z_i)$ in a point cloud $\mathbf{x}$ is encoded as 
$\mathrm{\Phi}(\mathbf{x})_i$ in some embedding space by a neural network encoder $\mathrm{\Phi}(\cdot)$, 
\eg~PointNet~\cite{pointnet} or PointNet++~\cite{qi2017pointnetpp}. 
Let the embeddings of a pair of points $p_i$ and $p_j$ from a shape be $\mathrm{\Phi}(\mathbf{x})_i$ and $\mathrm{\Phi}(\mathbf{x})_j$, 
normalized to unit length (\ie~$|| \mathrm{\Phi}(\mathbf{x})_i|| = 1$), and the set of convex components as 
described above be $\mathcal{C}$.
The pairwise loss is then defined as 

\begin{equation}
    \mathcal{L}^{pair}(\mathbf{x}, p_i, p_j, \mathcal{C}) =
        \begin{cases}
        1 -   \mathrm{\Phi}(\mathbf{x})_i^\top \mathrm{\Phi}(\mathbf{x})_j, & \text{if }  [\mathrm{\Gamma}(p_i, \mathcal{C}) = \mathrm{\Gamma}(p_j, \mathcal{C})]  \\
        \max(0, \mathrm{\Phi}(\mathbf{x})_i^\top \mathrm{\Phi}(\mathbf{x})_j - m), & \text{if } [\mathrm{\Gamma}(p_i, \mathcal{C}) \neq \mathrm{\Gamma}(p_j, \mathcal{C})].
        \end{cases}
    \label{eq:pairwise}
\end{equation}

This loss encourages
points belonging to the same component to have a high similarity $ \mathrm{\Phi}(\mathbf{x})_i^\top \mathrm{\Phi}(\mathbf{x})_j$,
and encourages points from different components to have low similarity, subject to a margin $m$ (set to $m=0.5$ as in \cite{kong2018recurrent}).
$[\cdot]$ denotes the Iverson bracket.

\vspace{2mm}
\noindent
\textbf{Joint training with ACD.}
Formally, let us consider samples $\mathcal{X} = \{ \mathbf{x}_i \}_{i \in [n]}$, 
divided into two parts: $\mathcal{X^L}$ and  $\mathcal{X^U}$ of sizes $l$ and $u$ 
respectively. Now $\mathcal{X^L} := \{\mathbf{x}_1, ... , \mathbf{x}_l  \}$ consist 
of point clouds that are provided with human-annotated 
labels $\mathcal{Y^L} := \{\mathbf{y}_1, ... , \mathbf{y}_l  \}$, 
while we do not know the labels of the samples $\mathcal{X^U} := \{\mathbf{x}_{l+1},
... , \mathbf{x}_{l+u}  \}$.
By running ACD on the samples in $\mathcal{X^U}$, we can obtain a 
set of components for each shape. The pairwise contrastive loss 
$\mathcal{L}^{pair}$ (Eq.~\ref{eq:pairwise}) can then be defined over $\mathbf{x}_i \in \mathcal{X^U}$ as a self-supervised objective.
For the samples $\mathbf{x}_i \in \mathcal{X^L}$, we have access to
their ground-truth labels $\mathcal{Y^L}$, which may for example, 
be semantic part labels. In that case, the standard choice of 
training objective is the \textit{cross-entropy loss} $\mathcal{L}^{CE}$, 
defined over the points in an input point cloud. 
Thus, we can train a network on both $\mathcal{X^L}$ and 
$\mathcal{X^U}$ via a \textbf{\textit{joint loss}} that combines both 
the supervised ($\mathcal{L}^{CE}$) and self-supervised ($\mathcal{L}^{pair}$)
objectives,

\begin{equation}
    \mathcal{L} = \mathcal{L}^{CE} + \lambda \cdot \mathcal{L}^{pair}.
    \label{eq:joint}
\end{equation}

The scalar hyper-parameter $\lambda$ controls the relative 
strength between the supervised and self-supervised training 
signals. 
In the \textbf{\textit{pretraining}} scenario, when we 
\textit{only} have the unlabeled dataset $\mathcal{X^U}$ available,
we can train a neural network purely on the ACD parts by 
optimizing the $\mathcal{L}^{pair}$ objective.

\section{Experiments}

We demonstrate the effectiveness of the ACD-based self-supervision across a range of experimental scenarios. 
For all the experiments in this section we use ACDs computed on all shapes from the ShapeNetCore data~\cite{Chang2015ShapeNetAI},
which contains 57,447 shapes across 55 categories.
The decomposition was computed using a concavity tolerance of $1.5 \times 10^{-3}$ and a volumetric
grid of resolution $128^3$.
All the other parameters are set to their default values according to a 
publicly available implementation\footnote{\url{https://github.com/kmammou/v-hacd}} of~\cite{vhacd}.
The resulting decompositions have an average of 17 parts per shape.
The ACD computation takes $1.6$s per shape on an Intel i7-2600 3.4GHz using 8 cores. 

\subsection{Shape classification on ModelNet}
In this set of experiments, we show that the representations learned by a network trained on ACD are useful for discriminative downstream tasks such as classifying point clouds into shape categories.

\vspace{2mm}
\noindent
\textbf{Dataset.} We report results on the ModelNet40 shape classification benchmark, which consists of 12,311 shapes from 40 shape categories in a train/test split of 9,843/2,468. A linear SVM is trained on the features extracted on the training set of ModelNet40. 
This setup mirrors other approaches for unsupervised learning on point clouds, such as FoldingNet~\cite{yang2018foldingnet} and Hassani~\etal~\cite{hassani2019unsupervised}.

\vspace{2mm}
\noindent
\textbf{Experimental setup.} 
A PointNet++ network is trained on the unlabeled ShapeNet-Core data using the pairwise contrastive loss on the ACD task, using the Adam optimizer, initial learning rate of $10^{-3}$ and halving the learning rate every epoch. 
This network architecture creates an embedding for each of the $N$ points in an input shape, 
while for the shape classification task we require a single global descriptor for the entire point cloud.
Therefore, we aggregate the per-point features of PointNet++ at the first two set aggregation layers (\texttt{SA1} and \texttt{SA2}) and the last fully connected layer (\texttt{fc}), resulting in 128, 256 and 128 dimensional feature vectors, respectively. Since features from different layers may have different scales, we normalize each vector to unit length before concatenating them, and apply element-wise signed square-rooting~\cite{sanchez2013image}, resulting in a final 512-dim descriptor for each point cloud.
The results are presented in Table~\ref{tab:modelnet}.
 
\vspace{2mm}
\noindent
\textbf{Comparison with baselines.}
As an initial na\"ive baseline, we use a PointNet++ network with random weights as our feature extractor, 
and then perform the usual SVM training. 
This gives 78\% accuracy on ModelNet40 -- while surprisingly good, the performance is not entirely unexpected: 
randomly initialized convolutional neural networks are known to provide useful features by virtue of their architecture, 
as studied in Saxe~\etal~\cite{saxe2011random}. 
Training this network with ACD, on the other hand, gives a significant boost to performance (78\% $\rightarrow$ \textbf{89.1}\%), 
demonstrating the effectiveness of our proposed self-supervision task. 
This indicates some degree of generalization across datasets and tasks -- from distinguishing convex components on ShapeNet to classifying shapes on ModelNet40.
Inspired by~\cite{hassani2019unsupervised}, we also investigated if adding a reconstruction component to the loss
would further improve accuracy.
Reconstruction is done by simply adding an AtlasNet~\cite{atlasnet} decoder to our model and using Chamfer distance
as reconstruction loss.
Without the reconstruction term (i.e. trained only to perform ACD using contrastive loss), our accuracy ($89.1\%$) is the same as the multi-task learning approach presented in~\cite{hassani2019unsupervised}.
After adding a reconstruction term, we achieve an improved accuracy of  $\mathbf{89.8\%}$.
On the other hand, having just reconstruction without ACD yields an accuracy of $86.2\%$.
This shows not only that ACD is a useful task when learning representations for shape classification, but that it can also be combined
with shape reconstruction to yield complementary benefits.

\vspace{2mm}
\noindent
\textbf{Comparison with previous work.} 
Approaches for \textit{unsupervised} or \textit{self-supervised} learning on point clouds are listed in the upper portion of Table~\ref{tab:modelnet}. Our method achieves \textbf{89.1\%} classification accuracy from purely using the ACD loss, which is met only by the unsupervised multi-task learning method of Hassani~\etal~\cite{hassani2019unsupervised} (adding a reconstruction loss to our method slightly improves over the state-of-the-art: 89.1\% $\rightarrow$ \textbf{89.8\%}). We note that our method merely adds a contrastive loss to a standard architecture (PointNet++), without requiring a custom architecture and multiple pretext tasks as in \cite{hassani2019unsupervised}, which uses clustering, pseudo-labeling and reconstruction.

\begin{table}[t]
\centering
\caption{\small{Unsupervised shape classification on the ModelNet40 dataset.
The representations learned in the intermediate layers by a network trained for the ACD task on ShapeNet data are general enough to be useful for discriminating between shape categories on ModelNet40. 
}}
\label{tab:modelnet}
\begin{tabular}{@{\extracolsep{5pt}}lc}
\toprule
 \textbf{Method}                           & \textbf{Accuracy (\%)}  \\
\midrule 
  VConv-DAE~\cite{sharma2016vconv}         & 75.5    \\
  3D-GAN~\cite{wu2016learning}            & 83.3    \\
  Latent-GAN~\cite{achlioptas2017representation}        & 85.7    \\
  MRTNet~\cite{mrt18}                                   & 86.4 \\
  PointFlow~\cite{yang2019pointflow}                    & 86.8 \\
  FoldingNet~\cite{yang2018foldingnet}        & 88.4   \\
  PointCapsNet~\cite{zhao20193d}              & 88.9 \\
  Multi-task~\cite{hassani2019unsupervised}   & 89.1 \\
\midrule
  Our baseline (with Random weights)           & 78.0 \\
  With reconstruction term only          & 86.2 \\
  Ours with ACD    & 89.1 \\
  Ours with ACD + Reconstruction    & \textbf{89.8} \\
\bottomrule
\end{tabular}
\end{table}

\subsection{Few-shot segmentation on ShapeNet}

\noindent
\textbf{Dataset.} We report results on the \textbf{ShapeNetSeg} part segmentation benchmark~\cite{Chang2015ShapeNetAI}, which is a subset of the ShapeNetCore database with manual annotations (train/val/test splits of 12,149/1,858/2,874). It consists of 16 man-made shape categories such as airplanes, chairs, and tables, with manually labeled semantic parts (50 in total), such as wings, tails, and engines for airplanes; legs, backs, and seats for chairs, and so on. Given a point cloud at test time, the goal is to assign each point its correct part label out of the 50 possible parts.  Few-shot learning tasks are typically described in terms of ``$n$-way $k$-shot'' -- the task is to discriminate among $n$ classes and $k$ samples per class are provided as training data. We modify this approach to our setup as follows -- we select $k$ samples from each of the $n=16$ shape categories as the labeled training data, while the task remains semantic part labeling over the 50 part categories.

\begin{table}[t]
\centering
\caption{\small{Few-shot segmentation on the ShapeNet dataset (\textit{class avg. IoU} over 5 rounds). The number of shots or samples per class is denoted by $k$ 
for each of the 16 ShapeNet categories used for supervised training. Jointly training with the ACD task reduces overfitting when labeled data is scarce, leading to significantly better performance over a purely supervised baseline.
}}
\label{tab:shapenet}
\begin{tabular}{@{\extracolsep{5pt}}lcccc}
\toprule
Samples/cls.    & \textbf{k=1}      & \textbf{k=3}      & \textbf{k=5}     & \textbf{k=10}   \\
\midrule 
  Baseline      & 53.15 $\pm$ 2.49  & 59.54 $\pm$ 1.49  & 68.14 $\pm$ 0.90 & 71.32 $\pm$ 0.52  \\
  w/ ACD        & 61.52 $\pm$ 2.19  & 69.33 $\pm$ 2.85  & 72.30 $\pm$ 1.80 & 74.12 $\pm$ 1.17  \\
\toprule
                & \textbf{k=20}     & \textbf{k=50}      & \textbf{k=100}      & \textbf{k=inf}   \\
\midrule
  Baseline      & 75.22 $\pm$ 0.82  &  78.79 $\pm$ 0.44  &  79.67 $\pm$ 0.33   & 81.40 $\pm$ 0.44  \\
  w/ ACD        & 76.19 $\pm$ 1.18  &  78.67 $\pm$ 0.72  &  78.76 $\pm$ 0.61   & 81.57 $\pm$ 0.68  \\
  
\bottomrule
\end{tabular}
\end{table}

\noindent
\textbf{Experimental setup.} 
 For this task, we perform \textit{joint training} with two losses -- the usual cross-entropy loss over labeled parts for the training samples from ShapeNetSeg, and an additional contrastive loss over the ACD components for the samples from ShapeNetCore (Eq.~\ref{eq:joint}), setting $\lambda = 10$. In our initial experiments, we found joint training to be more helpful than pre-training on ACD and then fine-tuning on the few-shot task (an empirical phenomenon also noted in \cite{xie2019self}), and thereafter consistently used joint training for the few-shot experiments. All overlapping point clouds between the human-annotated ShapeNetSeg and the unlabeled ShapeNetCore were removed from the self-supervised training set. The $(x, y, z)$ coordinates of the points in each point cloud are used an the input to the neural network; we do not include any additional information such as normals or category labels in these experiments.

\begin{figure}[t]
    \centering
    \includegraphics[width=\linewidth]{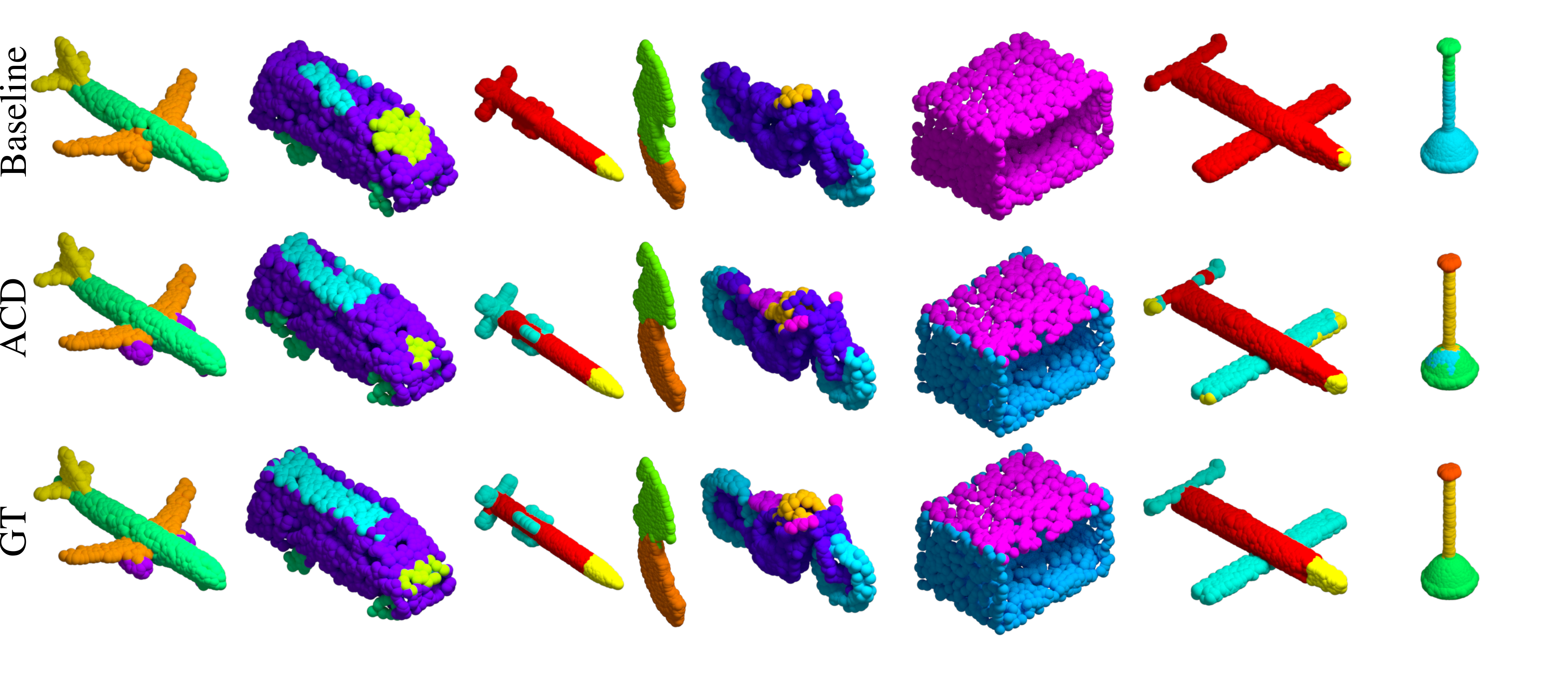}
    \caption{\small{Qualitative comparison on 5-shot ShapeNet~\cite{Chang2015ShapeNetAI} part segmentation.
    The baseline method in the first row corresponds to training using only 5 examples per class,
    whereas the ACD results in the second row were computed by performing joint training (cross-entropy from 5 examples + contrastive loss over ACD components from ShapeNetCore).
    The network backbone architecture is the same for both approaches -- PointNet++~\cite{qi2017pointnetpp}.
    The baseline method merges parts that should be separated, \eg, engines of the airplane,
    details of the rocket, top of the table, and seat of the motorcycle.}}
    \label{fig:segresults}
\vspace{-2mm}
\end{figure}

\vspace{2mm}
\noindent
\textbf{Comparison with baselines.}  Table~\ref{tab:shapenet} shows the few-shot segmentation performance of our method, versus a fully-supervised baseline. 
Especially in the cases of very few labeled training samples ($k=1, \dots, 10$), 
having the ACD loss over a large unlabeled dataset provides a consistent and significant gain 
in performance over purely training on the labeled samples. 
As larger amounts of labeled training samples are made available, 
naturally there is limited benefit from the additional self-supervised loss -- 
\eg~when using all the labeled data, our method is within standard deviation of the purely supervised baseline.
Qualitative results are shown in Fig.~\ref{fig:segresults}.

\vspace{2mm}
\noindent
\textbf{Comparison with previous work.}
The performance of recent \textit{unsupervised} and \textit{self-supervised} methods on ShapeNet segmentation are listed in Table~\ref{tab:sota_shapenet}. Consistent with the protocol followed by the multi-task learning approach of Hassani~\etal~\cite{hassani2019unsupervised}, we provide 1\% and 5\% of the training samples of ShapeNetSeg as the labeled data and report instance-average IoU. Our method clearly outperforms the state-of-the-art unsupervised learning approaches, improving over \cite{hassani2019unsupervised} at both the 1\% and 5\% settings (68.2 $\rightarrow$ \textbf{75.7}\% and 77.7 $\rightarrow$ \textbf{79.7}\%, respectively). 

\begin{table}[t]
\centering
\caption{\small{Comparison with state-of-the-art semi-supervised part segmentation methods on ShapeNet. Performance is evaluated using \textit{instance-averaged IoU}.
}}
\label{tab:sota_shapenet}
\begin{tabular}{@{\extracolsep{5pt}}lcc}
\toprule
 \multirow{2}{*}{\textbf{Method}}                          &  1\%~labeled         &  5\%~labeled \\
                                          & \textbf{IoU}         & \textbf{IoU} \\
\midrule 
 SO-Net~\cite{li2018so}                    &  64.0               & 69.0 \\
 PointCapsNet~\cite{zhao20193d}            &  67.0               & 70.0 \\
 MortonNet~\cite{MortonNet}               &  -               & 77.1 \\
 JointSSL~\cite{jointssl}\footnotemark                 &  71.9          & 77.4 \\
 Multi-task~\cite{hassani2019unsupervised} &  68.2               & 77.7 \\
\midrule
 ACD (\textit{ours})                       &  {\bf 75.7}    &  {\bf 79.7} \\ 
\bottomrule
\end{tabular}
\end{table}\footnotetext{Concurrent work.}

\subsection{Analysis of ACD}

\noindent
\textbf{Effect of backbone architectures.}
Differently from~\cite{chen2019bae,hassani2019unsupervised,yang2018foldingnet}, 
the ACD self-supervision does not require any custom network design and should be easily applicable across various backbone architectures. 
To this end, we use two recent high-performing models -- \textit{PointNet++} 
(with multi-scale grouping~\cite{qi2017pointnetpp}) and \textit{DGCNN}~\cite{wang2019dynamic} -- as the backbones, 
reporting results on ModelNet40 shape classification and few-shot segmentation ($k=5$) on ShapeNetSeg (Table~\ref{tab:arch_modelnet}). 
\begin{wrapfigure}{r}{0.4\textwidth}
    \centering
    \vspace{-0.5cm}
    \includegraphics[width=0.4\textwidth]{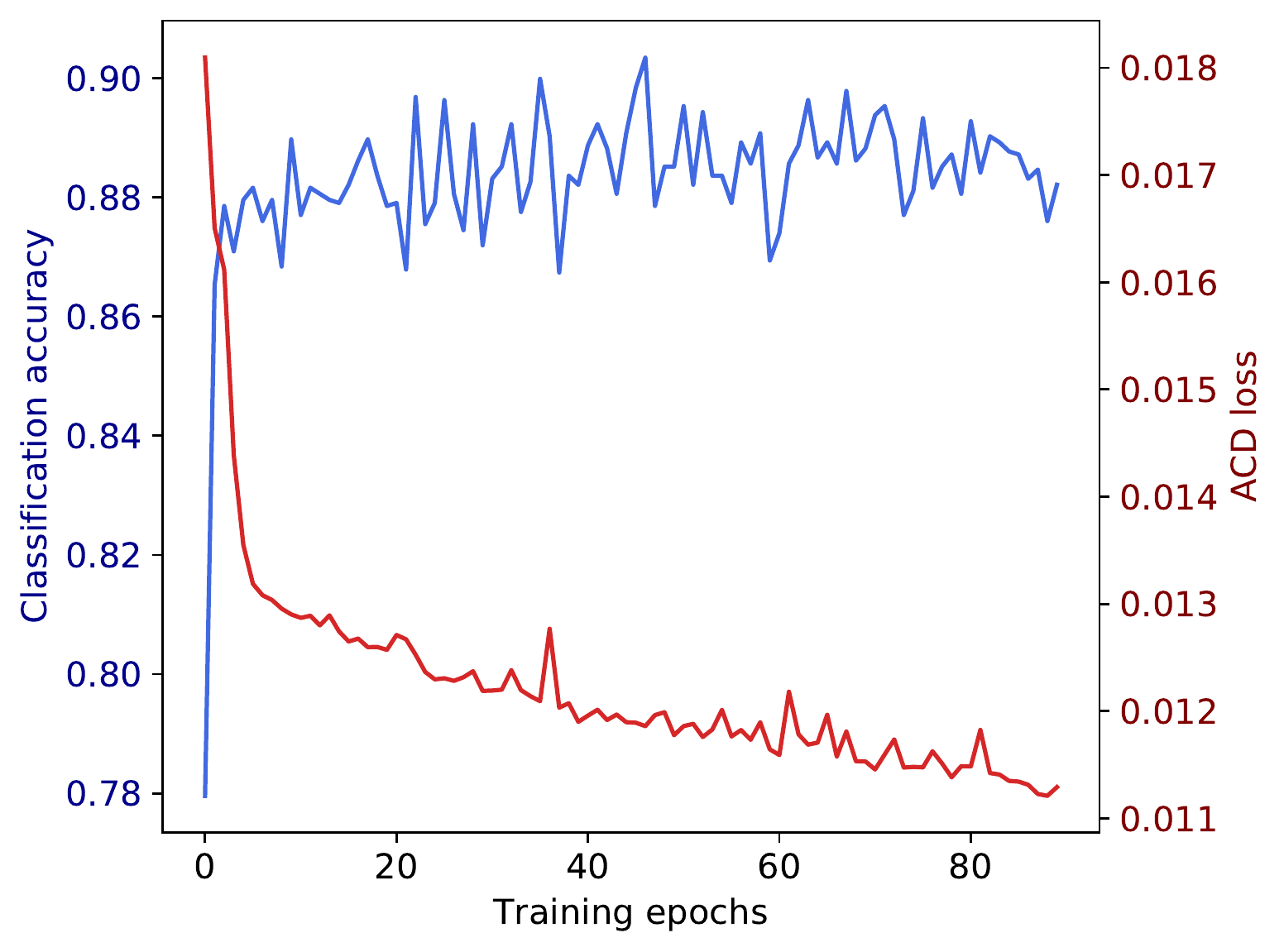}
    \vspace{-0.8cm}
    \caption{\small{Classification accuracy of a linear SVM on the ModelNet40 \emph{validation set} v.s. 
    the ACD \emph{validation loss} over training epochs. 
    }}
    \vspace{-0.5cm}
    \label{fig:acd_loss_svm}
\end{wrapfigure}
On shape classification, both networks show large gains from ACD pre-training: 11\% for PointNet++ (as reported earlier) and 14\% for DGCNN.
On few-shot segmentation with 5 samples per category (16 shape categories), PointNet++ improves from 68.14\% IoU to 72.3\% with the inclusion of the ACD loss. The baseline DGCNN performance with only 5 labeled samples per class is relatively lower (64.14\%), however with the additional ACD loss on unlabeled samples, the model achieves 73.11\% IoU, which is comparable to the corresponding PointNet++ performance (72.30\%). 

\begin{table}[t]
\centering
\caption{\small{Comparing embeddings from PointNet++~\cite{qi2017pointnetpp} and DGCNN~\cite{wang2019dynamic} backbones: shape classification accuracy on ModelNet40 (\textit{Class./MN40}) and few-shot part segmentation performance in terms of class-averaged IoU on ShapeNet (\textit{Part Seg./ShapeNet}).
}}
\label{tab:arch_modelnet}
\begin{tabular}{@{\extracolsep{6pt}}llcc}
\toprule
\textbf{Task / Dataset}      & \textbf{Method}   & \textbf{PointNet++}  & \textbf{DGCNN} \\
\midrule
\multirow{2}{*}{Class./MN40} & Baseline    & 77.96        &  74.11 \\
                             & w/ ACD  & \textbf{89.06}   &  \textbf{88.21} \\
\midrule
\multirow{2}{*}{5-shot Seg./ShapeNet} & Baseline   & 68.14 $\pm$ 0.90 & 64.14 $\pm$ 1.43 \\
                             & w/ ACD         & \textbf{72.30} $\pm$ 1.80 & \textbf{73.11} $\pm$ 0.95 \\
\bottomrule
\end{tabular}
\end{table}

\vspace{1mm}
\noindent
\textbf{Role of ACD in shape classification.} 
Fig.~\ref{fig:acd_loss_svm} shows the reduction in validation loss on learning ACD (red curve) as training progresses on the unlabeled ShapeNet data.  Note that doing well on ACD (in terms of the validation loss) also leads to learning representations that are useful for the downstream tasks of shape classification (in terms of SVM accuracy on a validation subset of ModelNet40 data, shown in blue).
However, the correlation between the two quantities is not very strong (Pearson $\rho = 0.667$) -- from the plots it appears that after the initial epochs, where we observe a large gain in classification accuracy as well as a large reduction in ACD loss, continuing to be better at the pretext task does not lead to any noticeable gains in the ability to classify shapes: training with ACD gives the model some useful notion of grouping and parts, but it is not intuitively obvious if \textit{perfectly} mimicking ACD will improve representations for classifying point-clouds into shape categories.

\begin{figure}[t]
    \centering
    \begin{tabular}{@{\extracolsep{1pt}}cccc}
    ACD & K-means & Spectral & HAC \\
    
    \includegraphics[width=0.23\textwidth]{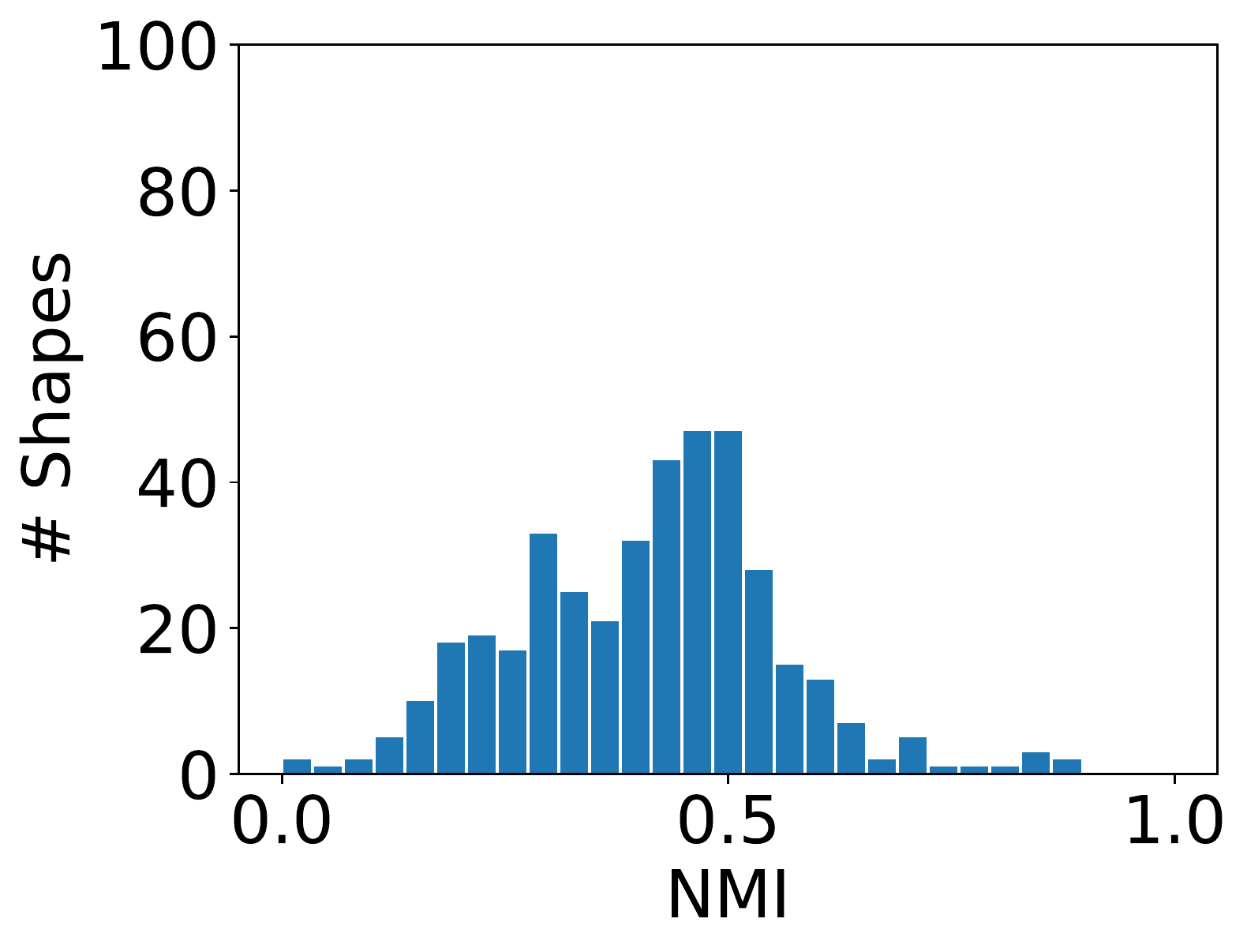} &
    \includegraphics[width=0.22\textwidth]{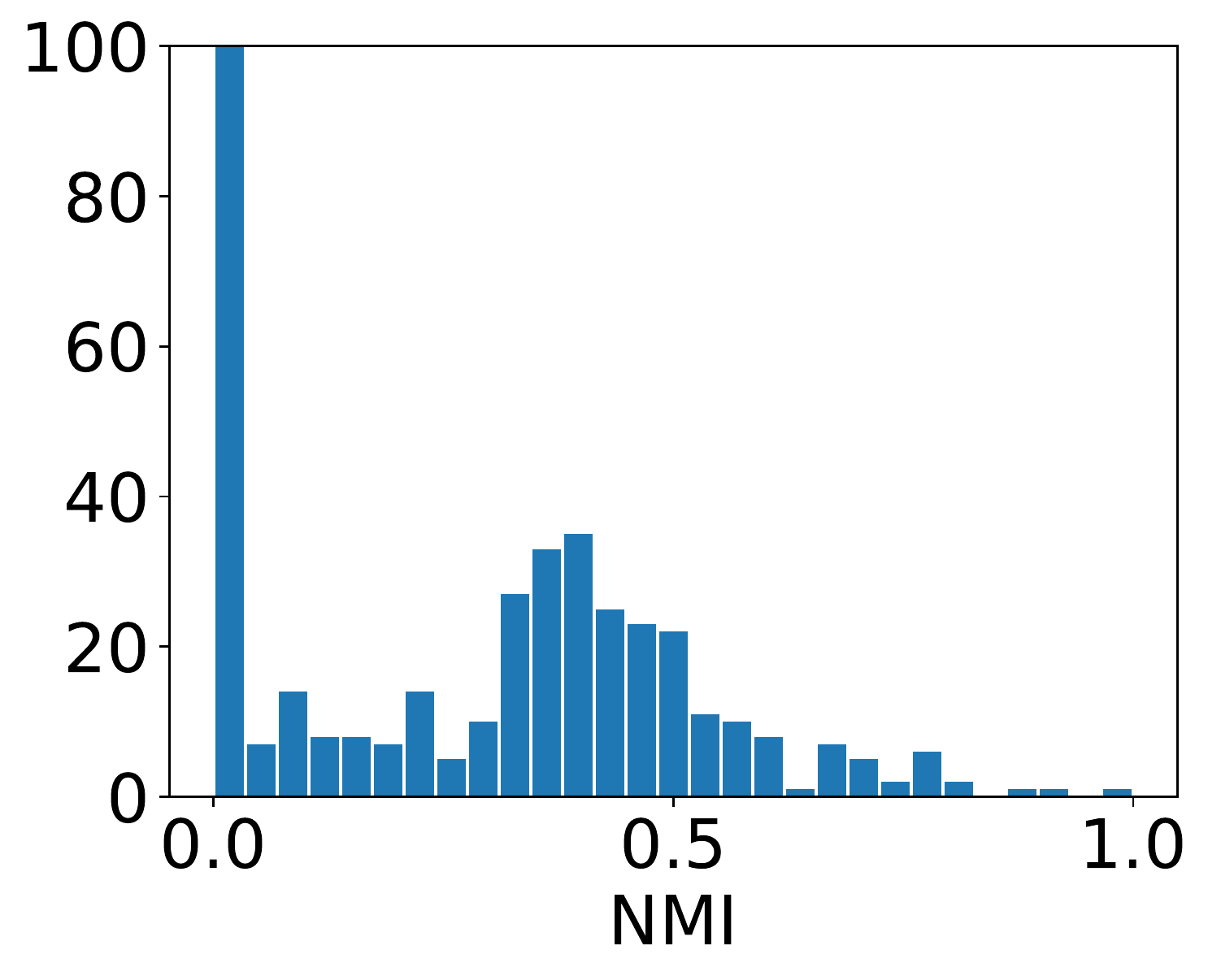} & 
    \includegraphics[width=0.22\textwidth]{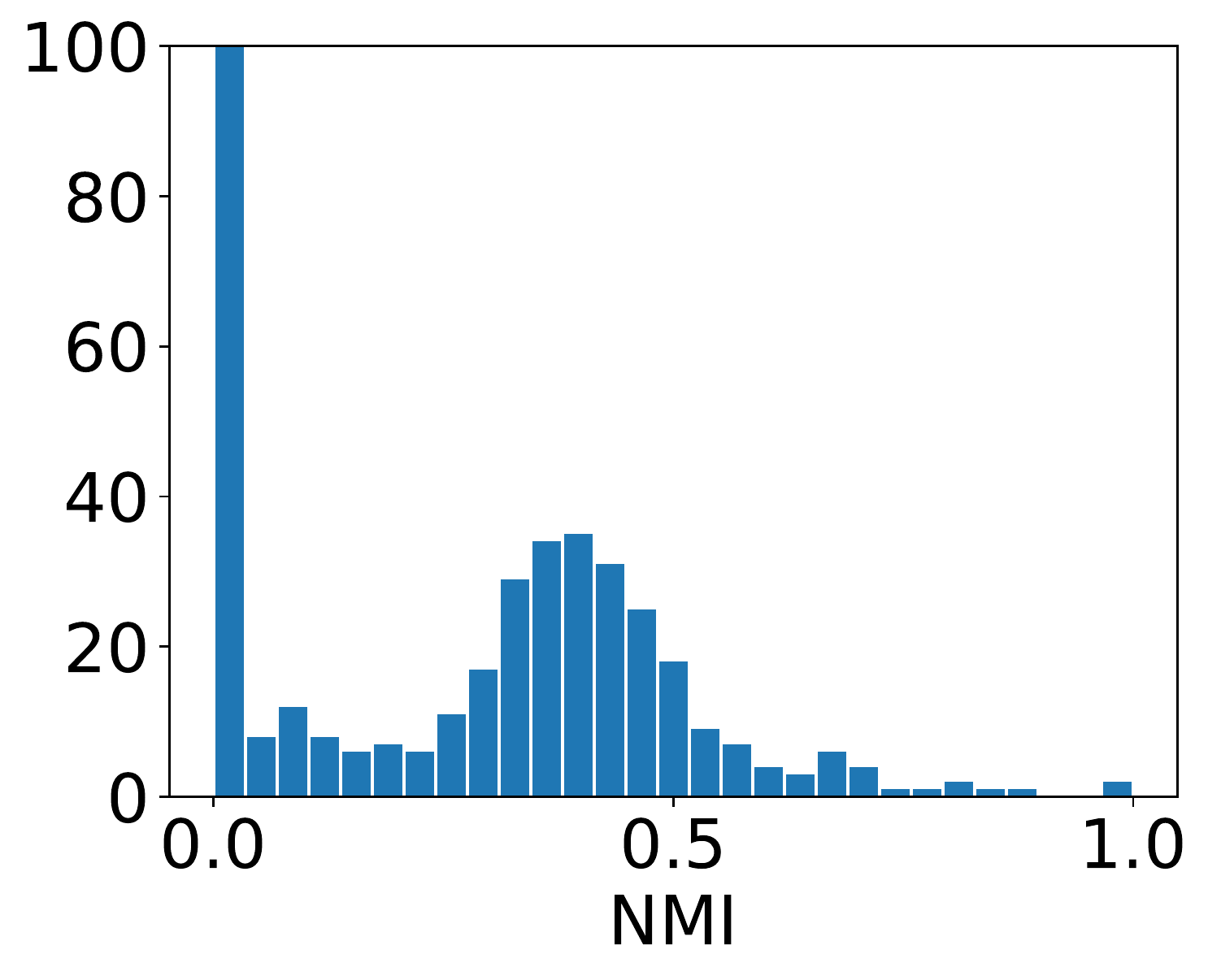} & 
    \includegraphics[width=0.22\textwidth]{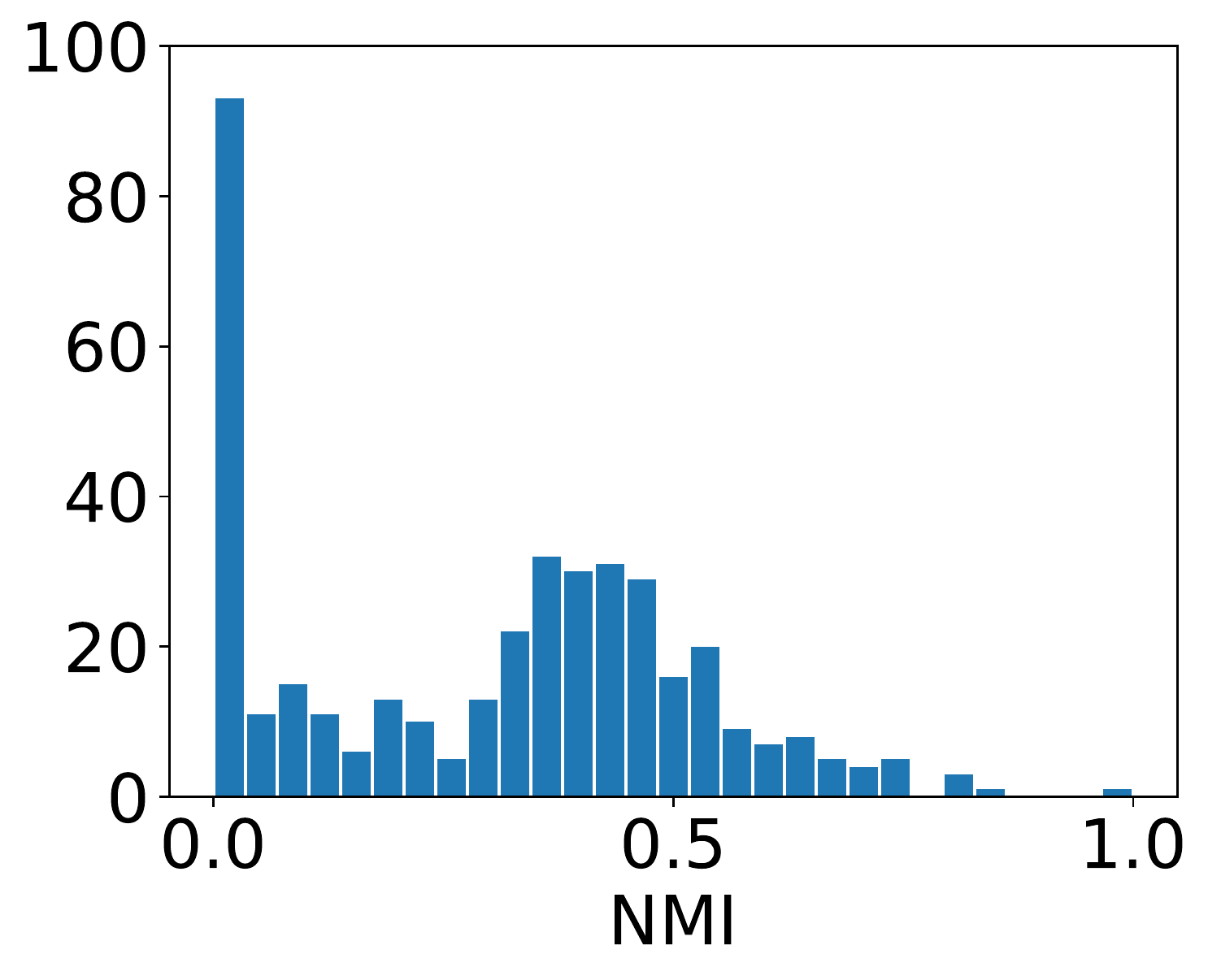} \\
    \hspace{0.1cm}
    \includegraphics[width=0.24\textwidth]{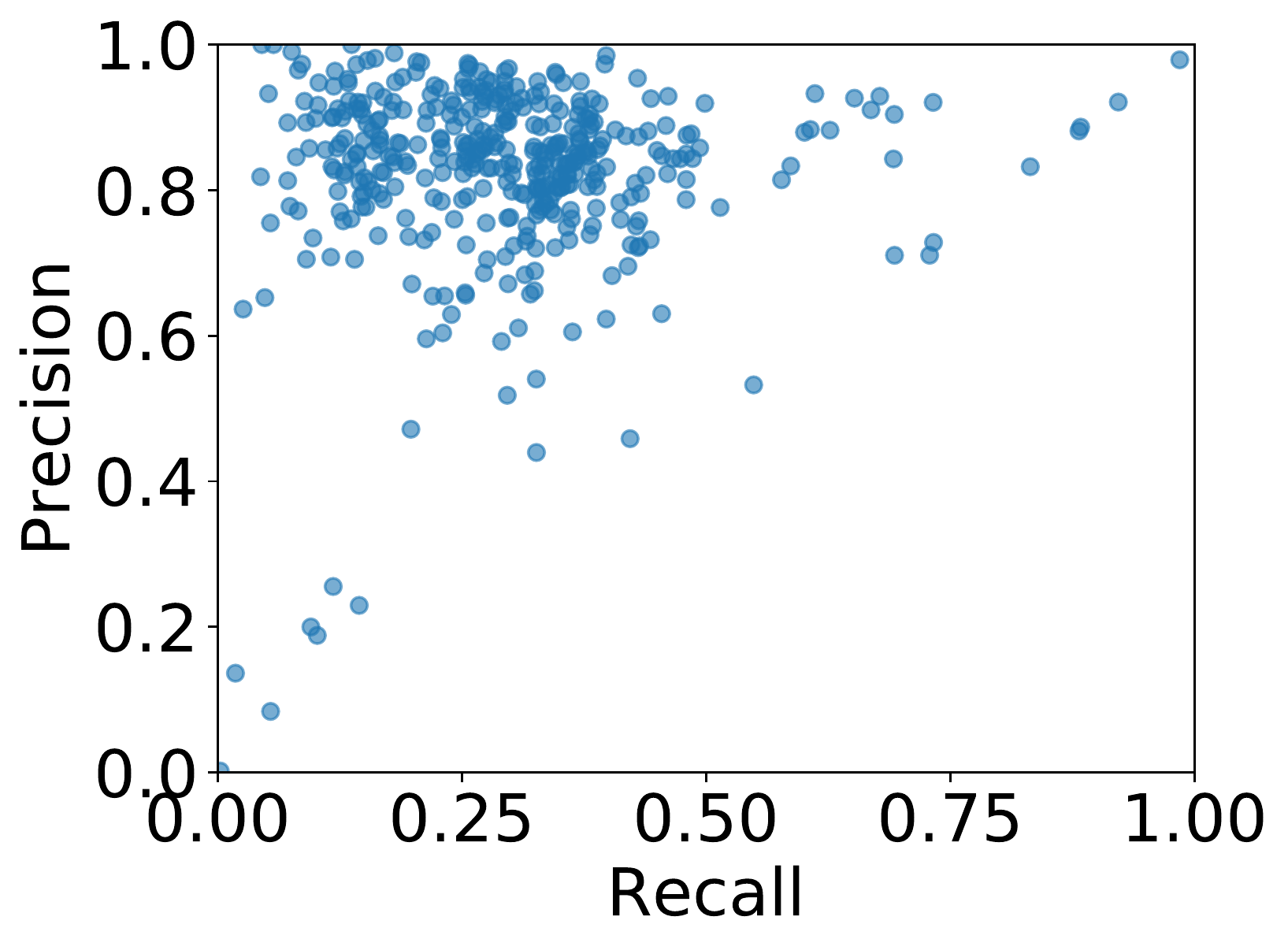} &
    \hspace{0.1cm}
    \includegraphics[width=0.22\textwidth]{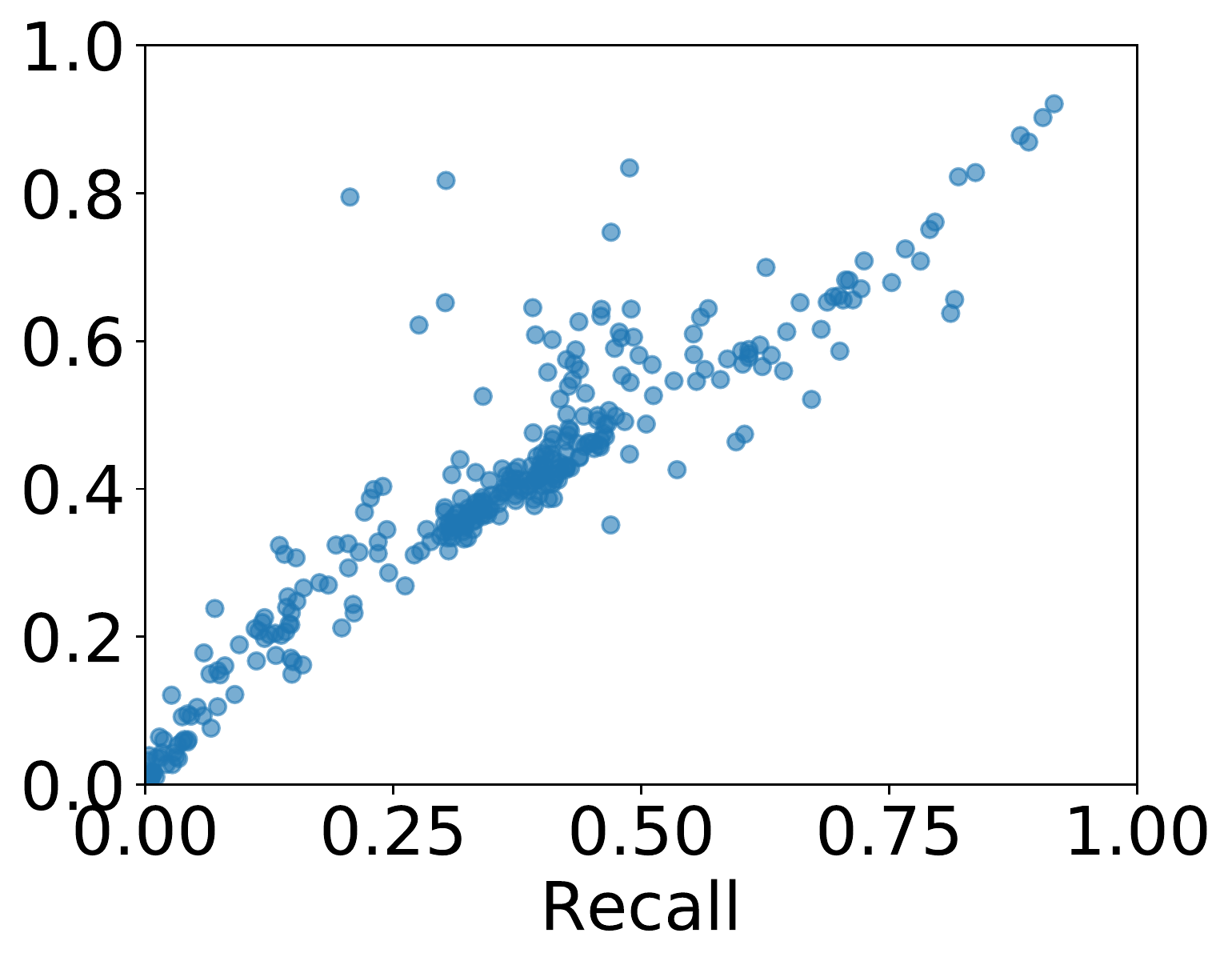} &
    \hspace{0.1cm}
    \includegraphics[width=0.22\textwidth]{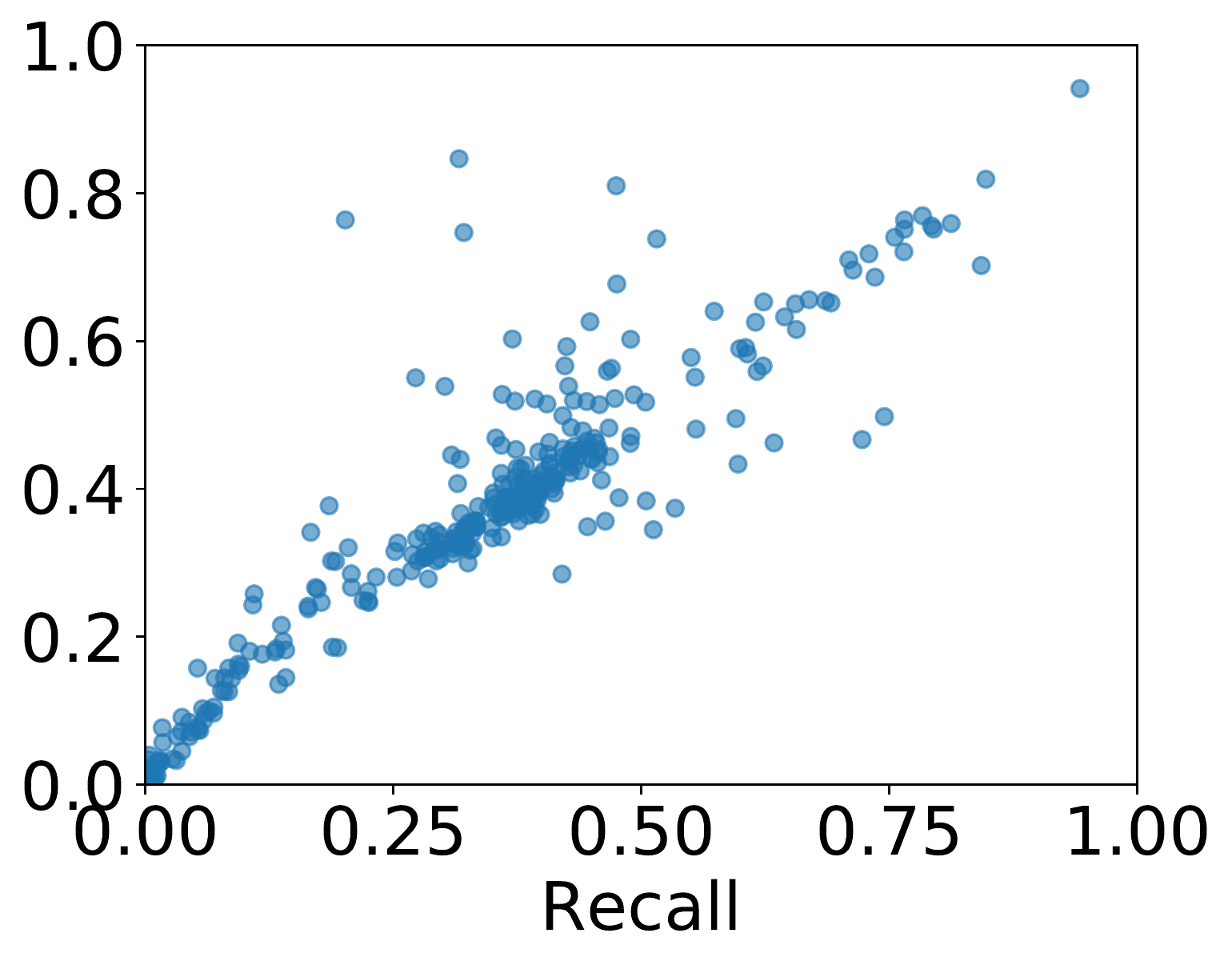} &
    \hspace{0.1cm}
    \includegraphics[width=0.22\textwidth]{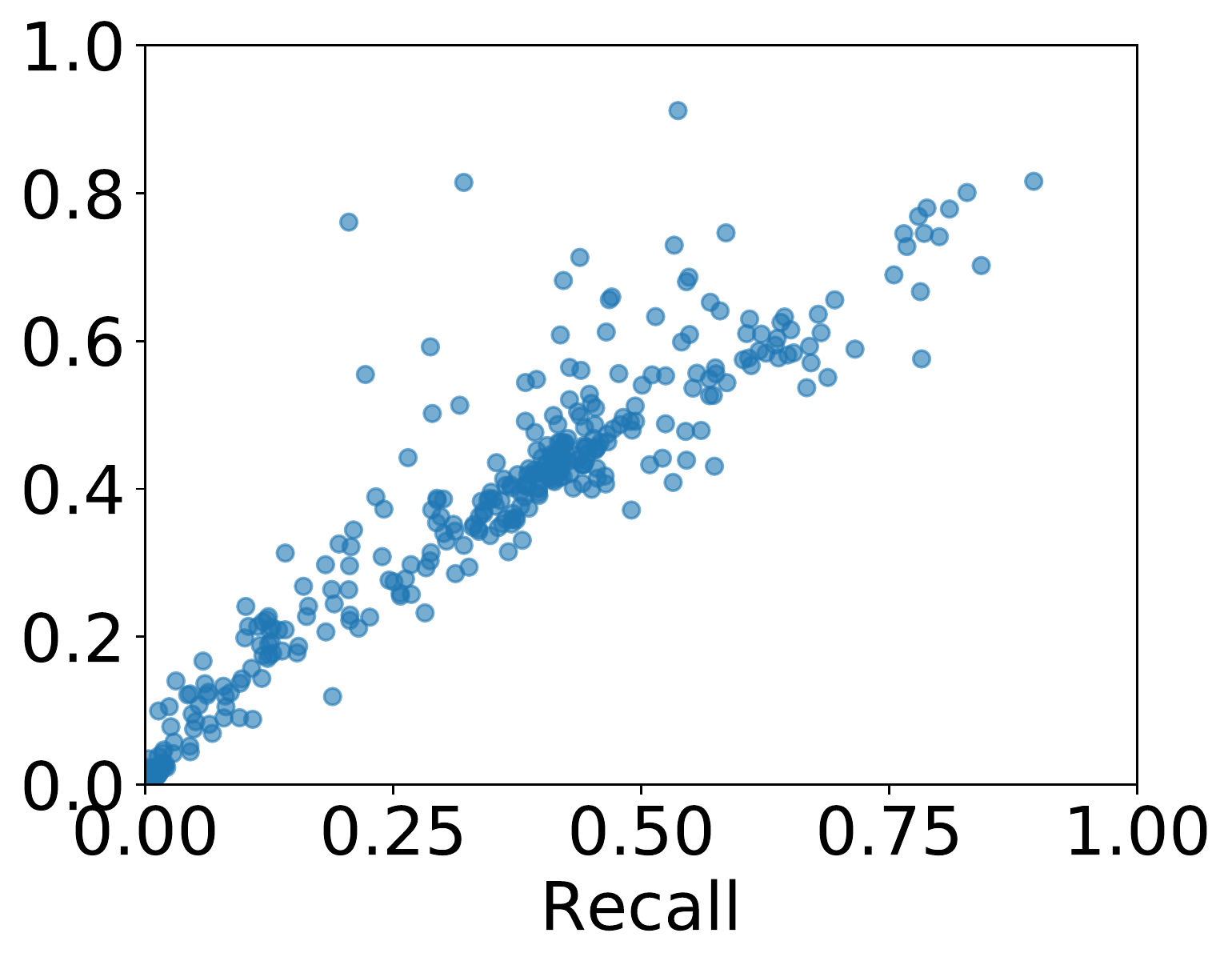} \\
    \end{tabular}
    \caption{ \small{Correspondence between human part labels and shape decompositions: comparing ACD with basic clustering algorithms -- K-means, spectral clustering and hierarchical agglomerative clustering (HAC). \textbf{\textit{Row-1:}} histogram of \textit{normalized mutual information} (NMI) between human labels and clustering -- ACD is closer to the ground-truth parts than others (y-axes clipped at 100 for clarity). 
    \textbf{\textit{Row-2:}} plotting \textit{precision v.s. recall} for each input shape, ACD has high precision and moderate recall (tendency to over-segment parts), while other methods are usually lower in both metrics.}}
    \label{fig:acd_nmi}
    \vspace{-0.5cm}
\end{figure}

\vspace{2mm}
\noindent
\textbf{Comparison with clustering algorithms.}
We quantitatively analyse the connection between convex decompositions and semantic object parts  by
comparing ACD with human part annotations on 400 shapes from ShapeNet, along with simple clustering baselines -- K-means~\cite{arthur2006k}, spectral clustering~\cite{shi2000normalized,von2007tutorial} 
and hierarchical agglomerative clustering (HAC)~\cite{mullner2013fastcluster} on $(x,y,z)$ coordinates 
of the point clouds. For the baselines, we set the number of clusters to be the number of ground-truth
parts in each shape. 
For each sample shape, given the set of $M$ part categories $\Omega = \{ \omega_1, \omega_2, \dots \omega_M \}$ 
and the set of $N$ clusters $\mathcal{C} = \{ C_1, C_2, \dots C_N  \}$, clustering performance is 
evaluated using \textit{normalized mutual information} (NMI)~\cite{vinh2010information}, defined as 
\begin{equation}
    \text{NMI}(\Omega, \mathcal{C}) = \frac{I(\Omega; \mathcal{C})}{ [H(\Omega) + H(\mathcal{C})]/ 2},
\end{equation}
where $I(\cdot ; \cdot)$ denotes the mutual information between classes $\Omega$ and clusters $\mathcal{C}$,
and $H(\cdot)$ is the entropy~\cite{cover2012elements}. A better clustering results in \textit{higher} NMI 
w.r.t. the ground-truth part labels. 
The first row of Fig.~\ref{fig:acd_nmi} shows the histograms of NMI between cluster assignments and human 
part annotations: ACD, though not
exactly aligned to human notions of parts, is significantly better than other clustering methods, which
have very low NMI in most cases. 

We also plot the \textit{precision} and \textit{recall} of clustering for each of the 400 shapes on the second row of
Fig.~\ref{fig:acd_nmi}. 
The other baseline methods show that a na\"ive clustering of points does not correspond well to semantic
parts. 
ACD has high precision and moderate recall on most of the shapes -- this agrees with the visual impression
that  the decompositions contain most of the boundaries present
in the human annotations, even if ACD tends to oversegment the shapes.
For example, ACD typically segments the legs of a chair into four separate components. On the other hand, the part annotations in
ShapeNet label all the legs of a chair with the same label, since the benchmark does not
distinguish between the individual legs of a chair.
We note that the correspondence of ACD to human part labels is not perfect, and this opens an interesting
avenue for further work -- exploring other decomposition methods like generalized cylinders~\cite{gdc}
that may correspond more closely to human-defined parts, and in turn could lead to improved downstream
performance on discriminative tasks.

\section{Conclusion}
Self-supervision using approximate convex decompositions (ACD) has
been shown to be effective across multiple tasks and datasets --
few-shot part segmentation on ShapeNet and shape classification on
ModelNet, consistently surpassing existing self-supervised and
unsupervised methods in performance.
A simple pairwise contrastive loss is sufficient for introducing the
ACD task into a network training framework, without dependencies on
any custom architectures.
The method can be easily integrated into existing state-of-the-art
architectures operating on point clouds such as PointNet++ and DGCNN,
yielding significant improvements in both cases.
Given the demonstrated effectiveness of ACD in self-supervision, this
opens the door to incorporating other shape decomposition methods from
the classical geometry processing literature into deep neural network models operating on point clouds.

\subsubsection*{Acknowledgements} The project is supported in part by the National Science Foundation (NSF) through grants \#1908669, \#1749833, \#1617333.  Our experiments were performed in the UMass GPU cluster obtained under the Collaborative Fund managed by the Massachusetts Technology Collaborative.

\bibliographystyle{splncs04}
\bibliography{egbib}
\end{document}